\documentclass{article}

 \usepackage[nonatbib,final]{neurips_2026}

\usepackage[utf8]{inputenc} 
\usepackage[T1]{fontenc}    
\usepackage{hyperref}       
\usepackage{url}            
\usepackage{booktabs}       
\usepackage{amsfonts}       
\usepackage{nicefrac}       
\usepackage{microtype}      
\usepackage[table]{xcolor}  
\usepackage{graphicx}       
\usepackage{amsmath}        
\usepackage{multirow}
\usepackage{tcolorbox}

\title{Imagine2Real: Towards Zero-shot Humanoid-Object Interaction via Video Generative Priors}

\definecolor{MutedBlue}{HTML}{4F71BE}
\definecolor{MutedOrange}{HTML}{DE8344}

\begin{document}
\begin{center}
    \LARGE 
    \textbf{
    \textcolor{MutedBlue!100!MutedOrange}{I}%
    \textcolor{MutedBlue!91!MutedOrange}{m}%
    \textcolor{MutedBlue!82!MutedOrange}{a}%
    \textcolor{MutedBlue!73!MutedOrange}{g}%
    \textcolor{MutedBlue!64!MutedOrange}{i}%
    \textcolor{MutedBlue!55!MutedOrange}{n}%
    \textcolor{MutedBlue!45!MutedOrange}{e}%
    \textcolor{MutedBlue!36!MutedOrange}{2}%
    \textcolor{MutedBlue!27!MutedOrange}{R}%
    \textcolor{MutedBlue!18!MutedOrange}{e}%
    \textcolor{MutedBlue!9!MutedOrange}{a}%
    \textcolor{MutedBlue!0!MutedOrange}{l}: Towards Zero-shot Humanoid-Object Interaction via Video Generative Priors
    }
\end{center}

\begin{center}
    Jiahe Chen$^{1,2\ast}$ \quad
    Zirui Wang$^{1,2\ast}$ \quad
    Feiyu Jia$^{2}$ \quad
    \\
    Xiao Chen$^{3,2}$ \quad
    Xiaojie Niu$^{2}$ \quad
    Weishuai Zeng$^{3,2}$ \quad
    \\
    Tianfan Xue$^{3}$ \quad
    Xiaowei Zhou$^{1}$ \quad
    Jiangmiao Pang$^{2\dagger}$ \quad
    Jingbo Wang$^{2\dagger}$
    \\
    \vspace{0.3cm}
    $^1$ Zhejiang University \quad
    $^2$ Shanghai AI Laboratory \quad
    $^3$ The Chinese University of Hong Kong \quad
    \vspace{0.1cm}
    \\
    \small{\textit{$\ast$: co-first authors  \quad $\dagger$: co-corresponding authors}}
\end{center}
\vspace{0.2cm}

\begin{figure}[h]
    \vspace{-0.4cm}
    \centering
    \includegraphics[width=\textwidth]{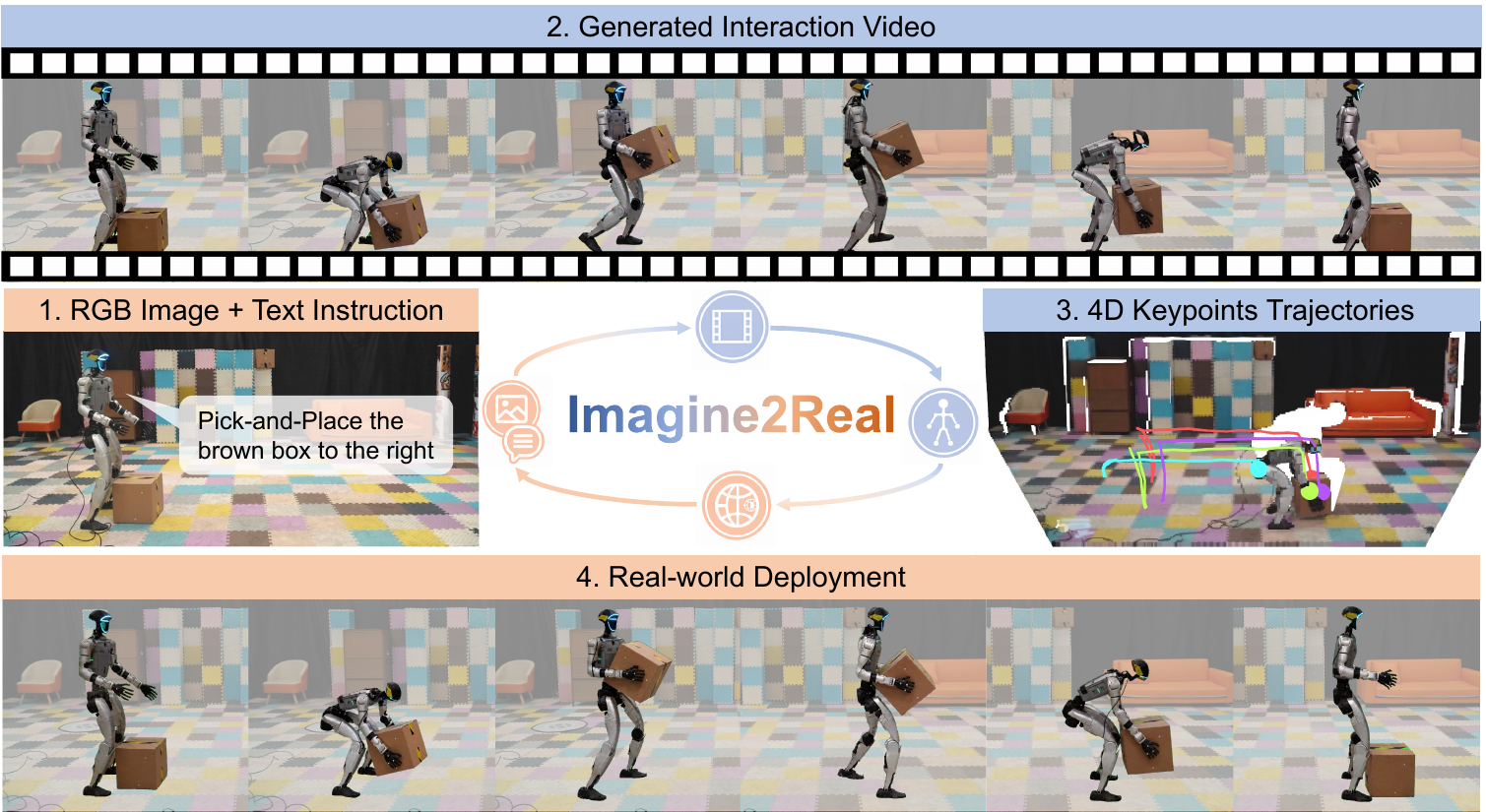}
    \caption{The Imagine2Real zero-shot deployment loop. Given (1) an image and text instruction, we (2) synthesize an interaction video, (3) extract 4D keypoints trajectories, and (4) deploy the policy within a mocap system, turning imagined interactions into physical reality.}
    \label{fig:teaser}
\end{figure}

\begin{abstract}
Whole-body Humanoid-Object Interaction (HOI) is bottlenecked by the scarcity of high-fidelity 3D data. While video generative priors offer a promising alternative, existing methods suffer from \textit{Representation Misalignment} due to their reliance on geometric priors (e.g., explicit CAD models), and \textit{Retargeting Complexity} arising from intensive morphing and morphological mismatch. We propose Imagine2Real, a zero-shot HOI framework for flexible, geometry-free interaction. To resolve misalignment, we formulate robot and object motions as unified 4D point trajectories. To overcome retargeting complexity, our Keypoints Tracker tracks only sparse critical points (base, hands, and object), entirely bypassing the error-amplifying retargeting process. To maintain natural gaits despite these sparse signals, we utilize the latent space of a Behavior Foundation Model (BFM) as the tracker's search domain. Using a progressive training strategy, Imagine2Real learns robust behaviors with simple tracking rewards, enabling zero-shot physical deployment within a motion capture(mocap) system.
\end{abstract}
\section{Introduction}
\label{sec:introduction}

Whole-body Humanoid-Object Interaction (HOI) is critical for deploying humanoid robots in unstructured environments such as households and factories. While proprioceptive control has matured significantly \cite{chen2025gmt, liao2025beyondmimic, luo2025sonic}, extending these capabilities to complex interactions remains difficult. The primary bottleneck is the lack of high-fidelity 3D interaction data for humanoid robots, which prevents the learning of robust and versatile interaction policies.

This data scarcity imposes a generalization ceiling on current HOI methodologies, where task versatility remains strictly limited by the available training samples. Scaling-based models such as VLA \cite{kim2024openvla, zitkovich2023rt, intelligence2026pi} pursue broad generalization through end-to-end training but are constrained by the lack of massive, humanoid-specific datasets. In contrast, approaches incorporating AMP \cite{peng2021amp, wang2025physhsi} utilize sparse, high-quality data to learn specialized behaviors, yet these methods often suffer from mode collapse\cite{salimans2016improved}, which restricts them to specific task categories and necessitates manual, task-specific reward engineering. Motion tracking methods \cite{zhao2025resmimic, weng2025hdmi} attempt to bypass data scarcity by mimicking single trajectories, but these are essentially limited to rigid motion replay and lack the autonomy needed for novel scenarios.

To address the shortage of 3D data, recent research employs internet-scale video generation models to provide visual and physical priors for downstream motion trackers. While this pipeline has shown promise in motion tracking \cite{ni2025generated}, applying it to HOI introduces two major challenges. 
\textit{1) Representation Misalignment.} Extracting 3D motion from 2D videos typically requires strong geometric priors (e.g., explicit object CAD models), which are seldom available in real-world settings. 
Furthermore, existing methods lack a unified way to jointly estimate robot-object motions. Estimating them independently through mature tools(e.g., GVHMR \cite{shen2024world} for robot and FoundationPose \cite{wen2024foundationpose} for object) introduces spatial and depth discrepancies, leading to low-quality reference motions and tracking failures. Consequently, many works resort to manually designing \cite{weng2025hdmi} or heuristically generating \cite{wang2026humanx} methods to rigidly attach objects to robot motions, which still requires CAD models and distorts actual object motion.
\textit{2) Retargeting Complexity}. Most motion tracking models rely on dense, low-level signals like joint positions. To obtain these from human videos, the extracted motion must be explicitly retargeted to a specific humanoid embodiment. In HOI scenarios, this retargeting becomes exceedingly complex: the introduction of object geometry requires intensive, time-consuming morphing \cite{yang2025omniretarget, ho2010spatial} to jointly align the robot and object meshes, which inherently amplifies estimation errors from the video source.
Avoiding this complex retargeting by tracking sparse keypoints is a promising alternative, but existing methods often struggle with quality. VisualMimic \cite{yin2025visualmimic} adopts 5-keypoints tracking but fails to achieve natural gaits due to the absence of joint-level information. HEAD \cite{chen2025hand} employs 3-keypoints tracking and relies on AMP to ensure natural locomotion; however, this reliance restricts its scalability from navigation to interaction tasks.

In this paper, we propose a zero-shot HOI framework that uses unified 3D point trajectories and a Keypoints Tracker based on the Behavior Foundation Model (BFM) \cite{li2025bfm, zeng2025behavior, luo2025sonic} to achieve flexible, geometry-free interaction. To resolve \textit{representation misalignment}, we extract robot and object motions using a unified 3D point tracker, representing both as 4D point trajectories. This shared representation avoids the need of geometric priors and ensures motions are aligned within a common frame. To address \textit{retargeting complexity}, our tracker follows only three interaction-critical points: the robot base and hands. By focusing on this minimal set, the framework entirely bypasses the complex, error-amplifying retargeting process. 

To achieve these capabilities despite scarce HOI data, we employ a three-stage progressive training strategy: (1) a whole-body tracking BFM on large-scale proprioceptive motions, (2) a keypoints tracker on loco-manipulation data, and (3) an interaction adaptor on specific HOI data. This approach ensures each stage benefits from the most relevant supervision at scale. By leveraging the BFM, sparse keypoints can still produce stable gaits, allowing the final stages to use simple tracking rewards instead of the complex machinery \cite{weng2025hdmi, zhao2025resmimic, xu2025intermimic} typically required for learning from scratch. 

This work makes three primary contributions. First, we introduce a unified 4D point trajectory representation for HOI that removes the dependency on geometric priors and resolves motion misalignment. Second, we develop a Keypoints Tracker based on the BFM that focuses on interaction-critical points, enabling flexible interaction alongside stable, natural gaits. Third, we design a progressive training framework that effectively utilizes multi-scale datasets to achieve robust zero-shot HOI without complex, hand-engineered mechanisms.
\section{Related Work}
\label{sec:related_work}

\subsection{Humanoid Behavior Foundation Models}
Humanoid motion tracking \cite{peng2018deepmimic, he2025asap} serves as the technical cornerstone for robust whole-body control, aiming to produce human-like movements by imitating reference trajectories. To scale these capabilities, recent research \cite{luo2023perpetual, chen2025gmt, he2024omnih2o, he2025hover, ji2024exbody2, ze2025twist2, zhu2026clot, pan2025agility, wang2026omnixtreme} focuses on training general-purpose motion trackers using large-scale datasets to reproduce diverse human behaviors.

Building upon these tracking techniques, Behavior Foundation Models (BFMs) \cite{li2025bfm, luo2025sonic, liao2025beyondmimic, yin2026unitracker, zeng2025behavior} have emerged to construct universal and reusable motion representation spaces. These models distill diverse motor skills into a compact latent space, providing strong physical priors for downstream tasks. Following this paradigm, our proposed framework trains a BFM backbone to function as a robust motion prior. By utilizing the learned latent representation of the BFM as the policy search space, the framework effectively constrains the movements of the humanoid, ensuring natural gaits and whole-body stability even in highly sparse keypoints tracking settings.

\subsection{Humanoid-Object Interaction}
Humanoid-Object Interaction (HOI) has recently achieved significant progress in complex tasks, such as ball sports \cite{zhang2026learning, ren2026smash, ren2025humanoid} and pick-and-place operations \cite{wang2025physhsi, su2025hitter, dao2024sim, liu2025opt2skill, zhang2024wococo}. However, many of these methods rely heavily on task-specific reward engineering, which limits their generalization across diverse scenarios.
Alternative research explores generalizable frameworks through the integration of multimodal instructions \cite{ze2025generalizable, fu2024humanplus, xue2025leverb, zhang2025falcon, ben2025homie}. Nevertheless, these approaches often isolate upper-body manipulation from lower-body locomotion or merely concatenate independent modules, thereby failing to exploit the intrinsic whole-body synergies and the high-DoF of the humanoid embodiment.
In contrast, recent works have successfully extended the motion tracking paradigm to interaction tasks, enabling humanoids to execute whole-body movements such as carrying boxes or pushing doors \cite{zhao2025resmimic, he2026ultra, weng2025hdmi, fu2025demohlm, yang2025omniretarget, liang2026interreal, lin2026pro, wang2026humanx}.

To enhance autonomy beyond trajectory replay, recent frameworks establish HOI generation and tracking pipelines \cite{yin2025visualmimic, kalaria2025dreamcontrol, allshire2025visual, yang2026zerowbc}. However, due to the scarcity of humanoid interaction data, these generators often synthesize only proprioceptive motions, restricting the robot to static environments or passive interactions. To address this data bottleneck, the proposed method utilizes video generation models as the upstream motion generator. By leveraging the rich visual and physical priors distilled from internet-scale video datasets, the framework achieves autonomous, high-quality, and general-purpose whole-body interactions.

\subsection{Video Generation for Robotics}
Recent advancements in video generation models have demonstrated remarkable capabilities in synthesizing realistic visual dynamics \cite{sora2024, gen3_2024, kling2024, wan2025, seedance2025}. Utilizing these priors, research in robotic manipulation tried to employ video generation as an intuitive world model \cite{patel2025robotic, ye2025textsc, hsieh2025dexman, liu2025geometry}. However, transferring this paradigm to humanoids presents unique challenges. Unlike fixed-base manipulators, humanoids must maintain dynamic stability during complex interactions despite severe data scarcity.

Pioneering this direction, GenMimic \cite{ni2025generated}  utilizes video generation to synthesize human motions and extract SMPL kinematics for whole-body tracking. However, as discussed in Section \ref{sec:introduction}, extending this pipeline to HOI encounters issues regarding \textit{Representation Misalignment} and \textit{Retargeting Complexity}. To bridge this gap, this work introduces unified 3D point trajectories coupled with a BFM-based Keypoints Tracker. By removing the dependency on geometric priors and bypassing complex retargeting, the framework successfully realizes a highly generalizable "video-to-motion-to-real" paradigm for HOI tasks.
\section{Methods}
\label{sec:methods}

\begin{figure}[t]
    \vspace{-0.3cm}
    \centering
    \includegraphics[width=\textwidth]{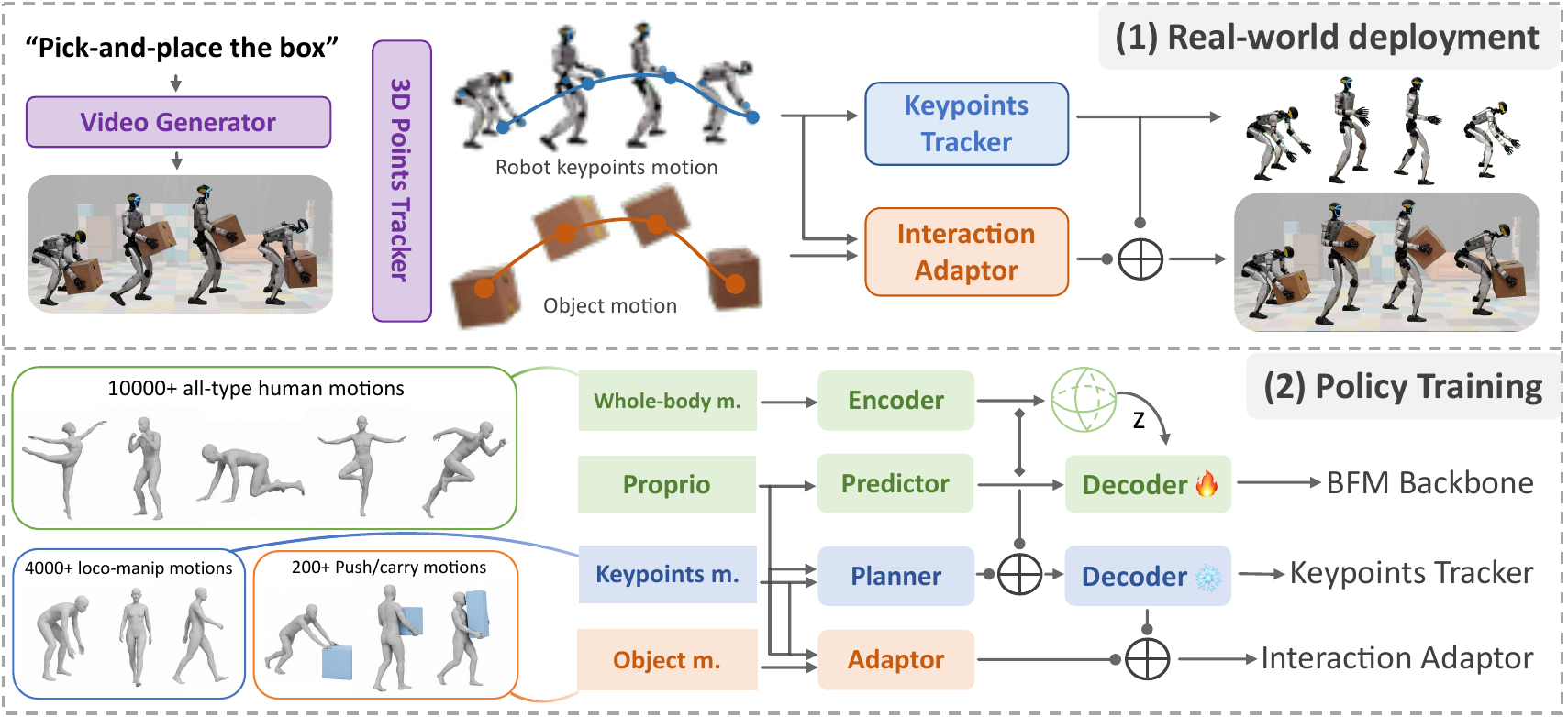}
    \caption{Overview of the Imagine2Real framework. \textbf{Top:} The zero-shot real-world deployment pipeline synthesizes an interaction video, extracts unified 3D point trajectories via a points tracker, and executes the motion using the Keypoints Tracker and Interaction Adaptor. \textbf{Bottom:} The policy training adopts a three-stage progressive strategy: (1) training a BFM backbone (Encoder, Predictor, Decoder) on diverse whole-body motions; (2) training a Keypoints Tracker (Planner) on sparse keypoints motions while keeping the BFM decoder frozen; and (3) training an Interaction Adaptor on object manipulation data to refine the final execution.}
    \label{fig:method}
    \vspace{-0.3cm}
\end{figure}

As illustrated in Figure~\ref{fig:method}, we address the zero-shot HOI task through a video-guided tracking paradigm. This involves synthesizing an interaction video, extracting reference motions, and controlling the robot to track them. Section \ref{subsec:training} details the progressive training strategy of the HOI Tracker, which sequentially trains a whole-body tracker backbone, a keypoints tracker, and an interaction adaptor to overcome the scarcity of interaction data. Section \ref{subsec:deployment} describes the real-world deployment pipeline, demonstrating the extraction of unified point trajectories from generated videos and their execution within a mocap system.

\subsection{HOI Tracker Training}
\label{subsec:training}

Given the disparity between abundant proprioceptive motion data and scarce interaction data, we design a progressive training framework to sequentially develop the humanoid's interaction capabilities. The pipeline consists of three cascaded modules. A whole-body motion tracker backbone is pre-trained on large-scale datasets (\textasciitilde 68.5h) to construct a compact motion representation space. A keypoints tracker is then trained on loco-manipulation data (\textasciitilde 8.86h) to track sparse points while maintaining balance. Finally, an interaction adaptor is fine-tuned on limited HOI data (\textasciitilde 0.43h). 
This progressive paradigm fully utilizes multi-level motion datasets under severe data scarcity. By decomposing the complex HOI task, the framework leverages abundant non-interactive data to build foundational priors, reducing the sample complexity required to learn interaction skills.

\subsubsection{BFM Backbone}
Directly learning natural sparse keypoints tracking from scratch via reinforcement learning is fundamentally difficult. When provided with only sparse tracking targets, a naive policy tends to aggressively overfit to the task objectives, severely violating essential physical constraints such as natural gaits and whole-body postural balance. To overcome this, we first train a Behavior Foundation Model (BFM) backbone to construct a versatile and reusable motion representation space. The BFM compresses local motion sequences into a shared latent motion space. At timestep $t$, given a $\delta$ steps motion subsequence $\mathbf{m}_{t:t+\delta}=[\mathbf{m}_{t}, \mathbf{m}_{t+1}, \dots, \mathbf{m}_{t+\delta}]$, a motion encoder $\mathcal{E}$ maps it to a latent code $\mathbf{z}_t = \mathcal{E}(\mathbf{m}_{t:t+\delta}) \in \mathcal{Z}$. Simultaneously, an autoregressive predictor $\mathcal{P}$ estimates a latent prior $\hat{\mathbf{z}}_t$ based solely on the proprioceptive observation history $\mathbf{o}_{0:t}^{\mathrm{prop}}$ and the previous latent state $\mathbf{z}_{t-1}$, formulated as $\hat{\mathbf{z}}_t = \mathcal{P}(\mathbf{o}_{0:t}^{\mathrm{prop}}, \mathbf{z}_{t-1})$.
To efficiently process the sequential input, the history $\mathbf{o}_{0:t}^{\mathrm{prop}}$ is recursively compressed into a compact hidden state using a Gated Recurrent Unit (GRU). The instantaneous observation at each time step is defined as $\mathbf{o}_t^{\mathrm{prop}} = [\mathbf{q}_t, \dot{\mathbf{q}}_t, \boldsymbol{\omega}_t, \boldsymbol{\psi}_t, \mathbf{p}_t, \mathbf{a}_{t-1}]$, capturing joint positions, velocities, base angular velocity, projected gravity, local end-effector positions relative to the base, and the previous action.
During the BFM pre-training, this predictor is optimized via a prediction loss to match the encoded reference motion, minimizing $\|\hat{\mathbf{z}}_t - \text{sg}(\mathbf{z}_t)\|_2^2$ (where $\text{sg}$ is the stop-gradient operator). The fundamental purpose of training this predictor is to internalize the forward dynamics and natural motion transitions of the humanoid. In downstream tasks where dense future reference motions are absent, the frozen predictor functions as an autonomous motor engine. It continuously outputs a physically viable, default motion prior $\hat{\mathbf{z}}_t$ to sustain natural locomotion and balance, which can then be slightly modulated by the higher-level sparse tracking policy.

Finally, a decoder $\mathcal{D}$ translates the current latent command $\mathbf{z}_t$ and proprioceptive history into the low-level action targets $\mathbf{a}_t = \mathcal{D}(\mathbf{o}_{0:t}^{\mathrm{prop}}, \mathbf{z}_t)$, where $\mathbf{a}_t$ is the action trained to track the dense whole-body motion. This pre-trained latent space $\mathcal{Z}$, along with the frozen predictor $\mathcal{P}$ and decoder $\mathcal{D}$, serves as a highly structured and physically grounded search space for keypoints tracking.

\subsubsection{Keypoints Tracker}

Rather than tracking the dense reference motion $\mathbf{m}_t$ of the whole body, the Keypoints Tracker is conditioned only on a sparse reference motion subsequence. This module does not relearn a new whole-body controller; instead, it performs task-conditioned biasing on top of the BFM low-level backbone. We treat the predictor $\mathcal{P}^*$ and decoder $\mathcal{D}^*$ as a frozen low-level backbone. Given the proprioceptive history and the previous latent command, the frozen predictor first produces a latent prior $\hat{\mathbf{z}}_t = \mathcal{P}^*(\mathbf{o}_{0:t}^{\mathrm{prop}}, \mathbf{z}'_{t-1})$. 

A latent residual planner $\mathcal{H}$ then uses the task commands (i.e., the sparse keypoints tracking targets $\mathbf{m}_{t:t+\delta}^{\mathrm{kp}}$), the previous latent $\mathbf{z}'_{t-1}$, and the current latent prior $\hat{\mathbf{z}}_t$ to output a residual command $\mathbf{z}^{\mathrm{res}}_t$. This residual is applied to the latent prior to form the new latent command $\mathbf{z}'_t$, which is then decoded into the full-body action $\mathbf{a}'_t$:
\begin{equation}
\begin{aligned}
\mathbf{z}^{\mathrm{res}}_t &= \mathcal{H}(\mathbf{o}_{0:t}^{\mathrm{prop}}, \mathbf{m}_{t:t+\delta}^{\mathrm{kp}}, \mathbf{z}'_{t-1}, \hat{\mathbf{z}}_t), \\
\mathbf{z}'_t &= \hat{\mathbf{z}}_t + \mathbf{z}^{\mathrm{res}}_t, \\
\mathbf{a}'_t &= \mathcal{D}^*(\mathbf{o}_{0:t}^{\mathrm{prop}}, \mathbf{z}'_t).
\end{aligned}
\end{equation}
Here, the superscript $*$ denotes that the corresponding modules are frozen during training. By applying low-dimensional latent residuals to the current latent prior, the policy preserves natural whole-body dynamics while tracking task-dependent sparse keypoint targets.

\subsubsection{Interaction Adaptor}

Since the BFM backbone is trained solely on non-interactive motion datasets, the latent space of the model inherently lacks the fine-grained dexterity required for object manipulation. To bridge this capability gap, we introduce an Interaction Adaptor that learns a residual joint-level action to refine the output of the Keypoints Tracker, drawing inspiration from the residual control technique used in ResMimic \cite{zhao2025resmimic}. Specifically, we formulate the Interaction Adaptor as a residual policy $\pi_{\mathrm{adapt}}$. At each time step, $\pi_{\mathrm{adapt}}$ receives the proprioceptive history $\mathbf{o}_{0:t}$, the sparse robot keypoints motion $\mathbf{m}_t^{\mathrm{kp}}$, and the target object motion $\mathbf{m}_t^{\mathrm{obj}}$. It then outputs a residual joint-level action $\Delta \mathbf{a}_t$, which is directly added to the output of the Keypoints Tracker $\mathbf{a}'_t$ to form the final execution command $\mathbf{a}^{\mathrm{int}}_t$:
\begin{equation}
\begin{aligned}
    \Delta \mathbf{a}_t &= \pi_{\mathrm{adapt}}(\mathbf{o}_{0:t}^{\mathrm{prop}}, \mathbf{m}_{t:t+\delta}^{\mathrm{kp}}, \mathbf{m}_{t:t+\delta}^{\mathrm{obj}}), \\
    \mathbf{a}^{\mathrm{int}}_t &= \mathbf{a}'_t + \Delta \mathbf{a}_t.
\end{aligned}
\end{equation}
This final adaptation enables precise, contact-rich interactions while preserving stable locomotion from previous stages. Guided by these foundational priors, the adaptor can be efficiently trained with straightforward tracking rewards, bypassing complex mechanisms typically required for learning from scratch, such as object force curriculum \cite{zhao2025resmimic}, contact rewards \cite{zhao2025resmimic, weng2025hdmi}, or Physical State Initialization (PSI) \cite{xu2025intermimic}.

\begin{figure}[t]
    \vspace{-0.3cm}
    \centering
    \includegraphics[width=\textwidth]{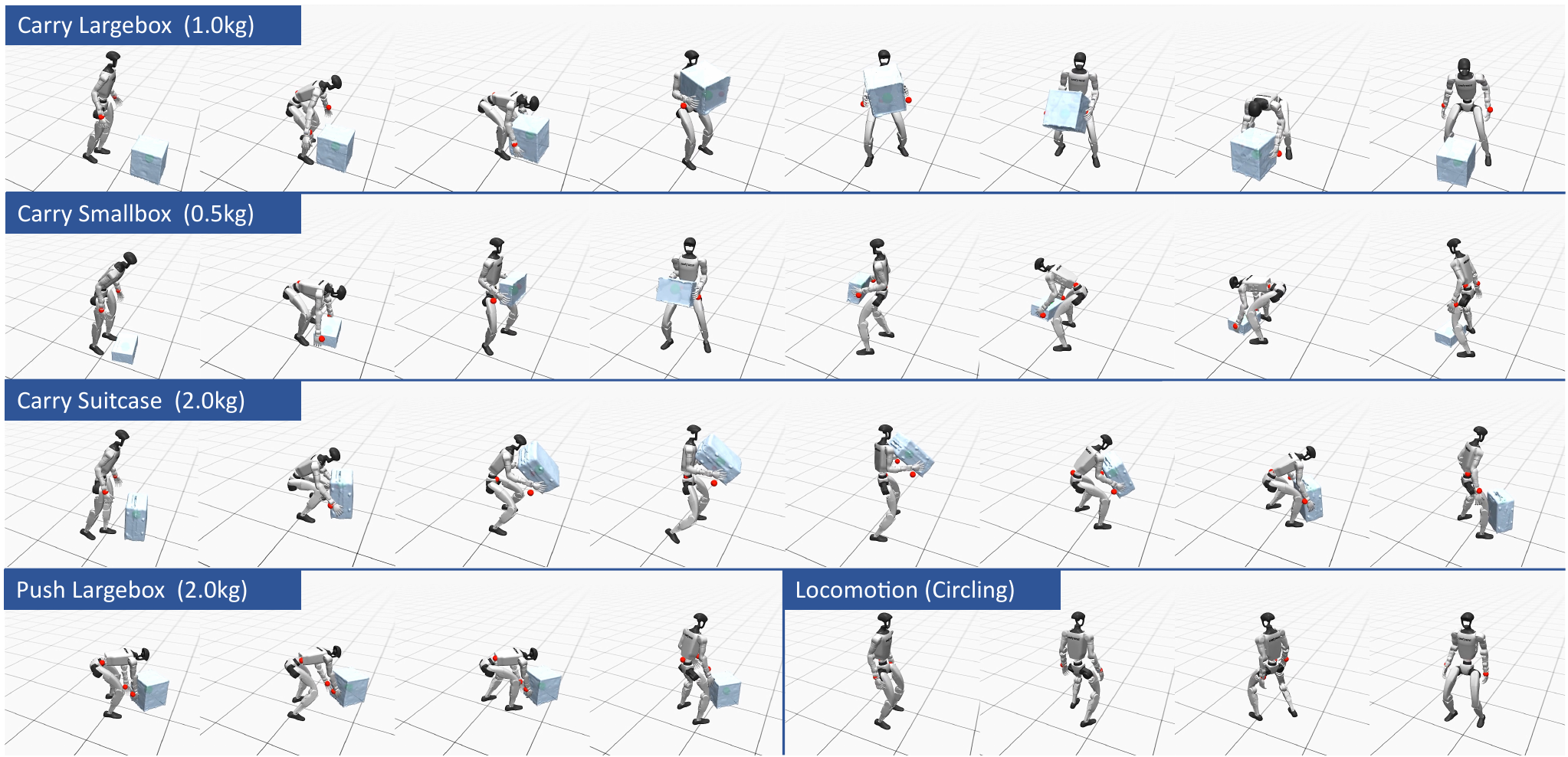}
    \caption{Qualitative results in simulation. Time-lapse sequences illustrate natural whole-body coordination across four HOI tasks (carrying/pushing boxes and carrying a suitcase), as well as the robust locomotion capability of our keypoints tracker.}
    \label{fig:result_sim}
    \vspace{-0.4cm}
\end{figure}

\subsection{Real-world Deployment}
\label{subsec:deployment}

\textbf{Video-to-Motion Pipeline.} Given a monocular RGB image containing the initial states of the robot and the target object, we synthesize an interaction video conditioned on a text instruction. From this video, we extract reference motions through a three-step process. First, we perform key region segmentation using SAM3 \cite{carion2025sam} to obtain precise masks for the robot's left hand, right hand, pelvis, and the target object. Second, we apply SpaTrackerV2 \cite{xiao2025spatialtrackerv2} to track the pixels within these masks, yielding 4D point trajectories for each region. Third, to address tracking noise, such as boundary points incorrectly classified as static background, we implement a robust keypoint filtering mechanism. We filter the tracked points based on the visibility and confidence scores output by SpaTrackerV2, subsequently removing spatio-temporal outliers. The remaining valid point trajectories within each mask are then geometrically averaged and smoothed to produce a single, stable reference trajectory for each corresponding keypoint.

\textbf{Mocap Deployment.} During real-world execution, we utilize a motion capture (mocap) system to fulfill two critical functions. First, it provides real-time global positions of the robot and object keypoints, which serve as inputs to the control policy for the tracking task. Second, it resolves the scale ambiguity inherent in video-generated motions. Because the synthesized videos lack metric depth information, the extracted trajectories represent only relative motions. We utilize the mocap system to directly capture the precise global positions of the keypoints at the initial frame and use them to scale and calibrate the generated trajectories to real-world dimensions. This approach yields higher calibration accuracy compared to monocular metric depth estimation or stereo vision. 

\section{Experiments}
\label{sec:experiments}

The experiments are designed to answer three core questions: (1) Does searching the BFM latent space yield more natural gaits and smoother motions than direct joint-space tracking? (2) For interaction, can our framework complete whole-body HOI tasks despite lacking dense full-body motion information such as joint positions? (3) For real-world deployment, can our pipeline successfully execute the zero-shot process—from video generation to physical HOI execution? 

\subsection{Experiment Setup}
\label{sec:experiment_setup}

\textbf{Dataset. } We construct a progressive training and data utilization framework to train the cascaded modules. (1) For the BFM-backbone, we utilize a massive corpus of human motion data from AMASS \cite{mahmood2019amass}, LAFAN1 \cite{harvey2020robust}, and 100STYLE \cite{mason2022real}, totaling over 10,000 motion clips(\textasciitilde 68.5h). These motions are retargeted to the robot using inverse kinematics via pink \cite{pink}. (2) For the Keypoints Tracker, we employ the complete body motions from LAFAN1 and OMOMO \cite{li2023object}, alongside selected loco-manipulation motions from AMASS. We filter out actions that are inherently difficult to represent with only three critical points (e.g., dancing and flipping), yielding over 4,000 motion clips(\textasciitilde 8.86h). (3) For the Interaction Adaptor, we select specific box-carrying and pushing tasks from OMOMO that do not require dexterous hand manipulation. We apply HOI retargeting via holosoma \cite{Amazon_FAR_and_Abbeel_Holosoma}, resulting in over 200 motion clips(\textasciitilde 0.43h). Note that the policy requires no retargeted data during inference; we use it during training solely to correct human-robot morphological mismatches and provide some auxiliary tracking rewards. 

\textbf{Metrics. } We evaluate our method across two primary dimensions: task completion and motion naturalness. \underline{(1) Task Completion:} \textit{Success Rate (\textbf{SR})} is the percentage of episodes where the relevant tracking errors (hands/base for tracking, object for interaction) remain below $0.2$\,m at all times. \textit{Tracking Errors ($\mathbf{E}_{\text{hands}}$, $\mathbf{E}_{\text{base}}$, $\mathbf{E}_{\text{obj}}$)} and their final-frame counterparts ($\mathbf{E}_{\text{hands-f}}$, $\mathbf{E}_{\text{base-f}}$, $\mathbf{E}_{\text{obj-f}}$) measure the Euclidean distance between generated and reference trajectories. \underline{(2) Motion Naturalness:} \textit{Mean Per Joint Angle Error ($\mathbf{E}_{\text{mpjae}}$)} reflects gait naturalness (computed against the ground-truth dense motion from which the sparse keypoints were extracted). \textit{Action Rate ($\mathbf{A}_{\text{rate}}$)} and \textit{Action Smoothness ($\mathbf{A}_{\text{smooth}}$)} compute the first- and second-order derivatives of actions, where lower values indicate smoother commands.

\textbf{Training Details. } We train all policies in Isaac Gym \cite{makoviychuk2021isaac} using PPO \cite{schulman2017proximal}. The training is accelerated with 8,192 parallel environments on a single RTX 4090 GPU. We employ the Adam optimizer with an initial learning rate of $3 \times 10^{-4}$. To simulate 3D tracker noise and enhance robustness, we inject 5cm Gaussian noise into the reference keypoints during the latter two training stages. Due to space constraints, comprehensive training details, including network architectures, hyperparameters, and reward formulations, are deferred to Appendix \ref{sec:appendix_training}. To ensure reproducibility, the anonymized source code and pre-trained checkpoints are provided in the supplementary material.

\begin{table}[t]
    \centering
    \caption{Quantitative evaluation of HOI task execution. We compare our full method (w/ Adaptor) against the baseline keypoints tracker (w/o Adaptor). We report the Success Rate (\textbf{SR}) and tracking errors (in cm) for the object ($\mathbf{E}_{\text{obj}}$), hands ($\mathbf{E}_{\text{hands}}$), and robot base ($\mathbf{E}_{\text{base}}$). The subscript ``-f'' denotes the final tracking error at the end of the episode.}
    \label{tab:result_task}
    \begin{tabular}{llccccccc}
    \toprule
    \textbf{Task} & \textbf{Method} & \textbf{SR} (\%) $\uparrow$ & $\mathbf{E}_{\text{obj}}$ $\downarrow$ & $\mathbf{E}_{\text{obj-f}}$ $\downarrow$ & $\mathbf{E}_{\text{hands}}$ & $\mathbf{E}_{\text{hands-f}}$ & $\mathbf{E}_{\text{base}}$ & $\mathbf{E}_{\text{base-f}}$ \\ 
    \midrule
    \multirow{2}{*}{Carry Box} & w/o Adaptor & 0.00 & - & - & - & - & - & - \\
    & w/ Adaptor & \cellcolor{green!10}82.65 & \cellcolor{green!10}6.34 & \cellcolor{green!10}7.03 & 7.33 & 4.46 & 5.16 & 4.05 \\
    \midrule
    \multirow{2}{*}{Push Box} & w/o Adaptor & 29.82 & 11.11 & 17.87 & 4.42 & 2.39 & 5.85 & 4.58 \\
    & w/ Adaptor & \cellcolor{green!10}64.91 & \cellcolor{green!10}9.00 & \cellcolor{green!10}13.02 & 7.25 & 4.81 & 6.49 & 4.40 \\
    \bottomrule
    \end{tabular}
\end{table}
\begin{table}[t]
    \centering
    \caption{Ablation study of the keypoints tracker. We compare our BFM-based approach against Direct tracking and DAgger baselines. The evaluation reports task success rate (\textbf{SR}), tracking accuracy ($\mathbf{E}_{\text{hands}}$, $\mathbf{E}_{\text{base}}$), motion naturalness metrics ($\mathbf{E}_{\text{mpjae}}$) and action smoothness metrics ($\mathbf{A}_{\text{rate}}$, $\mathbf{A}_{\text{smooth}}$).}
    \label{tab:ablation_tracker}
    \begin{tabular}{lcccccc}
        \toprule
        Method & \textbf{SR} (\%) $\uparrow$ & $\mathbf{E}_{\text{hands}}$ (cm) $\downarrow$ & $\mathbf{E}_{\text{base}}$ (cm) $\downarrow$ & $\mathbf{E}_{\text{mpjae}}$ (rad) $\downarrow$ & $\mathbf{A}_{\text{rate}}$ $\downarrow$ & $\mathbf{A}_{\text{smooth}}$ $\downarrow$ \\
        \midrule
        Direct     & 99.16 & \cellcolor{green!10}1.95 & \cellcolor{green!10}1.66 & 0.44 & 1.65 & 0.64 \\
        DAgger     & 99.32 & 3.73 & 3.32 & 0.36 & 0.61 & 0.20  \\
        Ours (BFM) & \cellcolor{green!10}99.36 & 3.08 & 3.78 & \cellcolor{green!10}0.25 & \cellcolor{green!10}0.22 & \cellcolor{green!10}0.09  \\
        \bottomrule
    \end{tabular}
    \vspace{-0.3cm}
\end{table}

\subsection{Sim2sim Evaluation}
\label{sec:sim_eval}

To verify policy generalization and ensure fair evaluation by preventing overfitting to the Isaac Gym environment, we conduct Sim2Sim evaluations in MuJoCo \cite{todorov2012mujoco}. All quantitative metrics reported in this section are evaluated in the MuJoCo simulator. 

\textbf{Ablation on Keypoints Tracking. } To answer the first question, we ablate our BFM-based Keypoints Tracker against two baselines: a Direct tracking approach (which uses the sparse points directly as commands to track in joint space) and a DAgger baseline (which distills a privileged whole-body tracker into a three-point tracking policy). As shown in \autoref{tab:ablation_tracker}, the Direct baseline achieves the lowest $\mathbf{E}_{\text{hands}}$ and $\mathbf{E}_{\text{base}}$ because it aggressively optimizes for point matching without any physical constraints. However, this comes at the cost of extreme jittering ($\mathbf{A}_{\text{rate}}=1.65$, $\mathbf{A}_{\text{smooth}}=0.64$) and unnatural whole-body postures ($\mathbf{E}_{\text{mpjae}}$ is exceptionally high), making it nearly impossible to deploy on a real robot. Distilling a prior via DAgger slightly mitigates these issues, but the errors remain relatively high. In contrast, by restricting the search space to the latent space of the BFM rather than the raw joint space, our framework significantly reduces $\mathbf{A}_{\text{rate}}$, $\mathbf{A}_{\text{smooth}}$, and $\mathbf{E}_{\text{mpjae}}$, proving that our method can execute the necessary tracking while maintaining relative natural gaits and soft actions.

\textbf{HOI Task Execution. } To answer the second question regarding interaction capability under sparse motion information, we evaluate our framework on carrying and pushing boxes. As detailed in \autoref{tab:result_task}, our framework successfully completes these tasks with high success rates. Furthermore, the comparison between the \textit{w/ Adaptor} and \textit{w/o Adaptor} settings demonstrates that the Interaction Adaptor is essential for bridging the gap between kinematic tracking and successful physical object manipulation.
The relatively lower success rate for pushing the large box stems from the lack of marker orientation input—omitted because it is notoriously difficult to robustly extract from videos, which hinders the robot from finely adjusting its hands during pushing. 
Qualitative results of these simulated interactions are visualized in \autoref{fig:result_sim}.

\subsection{Real-world Evaluation}
\label{sec:real_eval}

\begin{figure}[t]
    \vspace{-0.3cm}
    \centering
    \includegraphics[width=\textwidth]{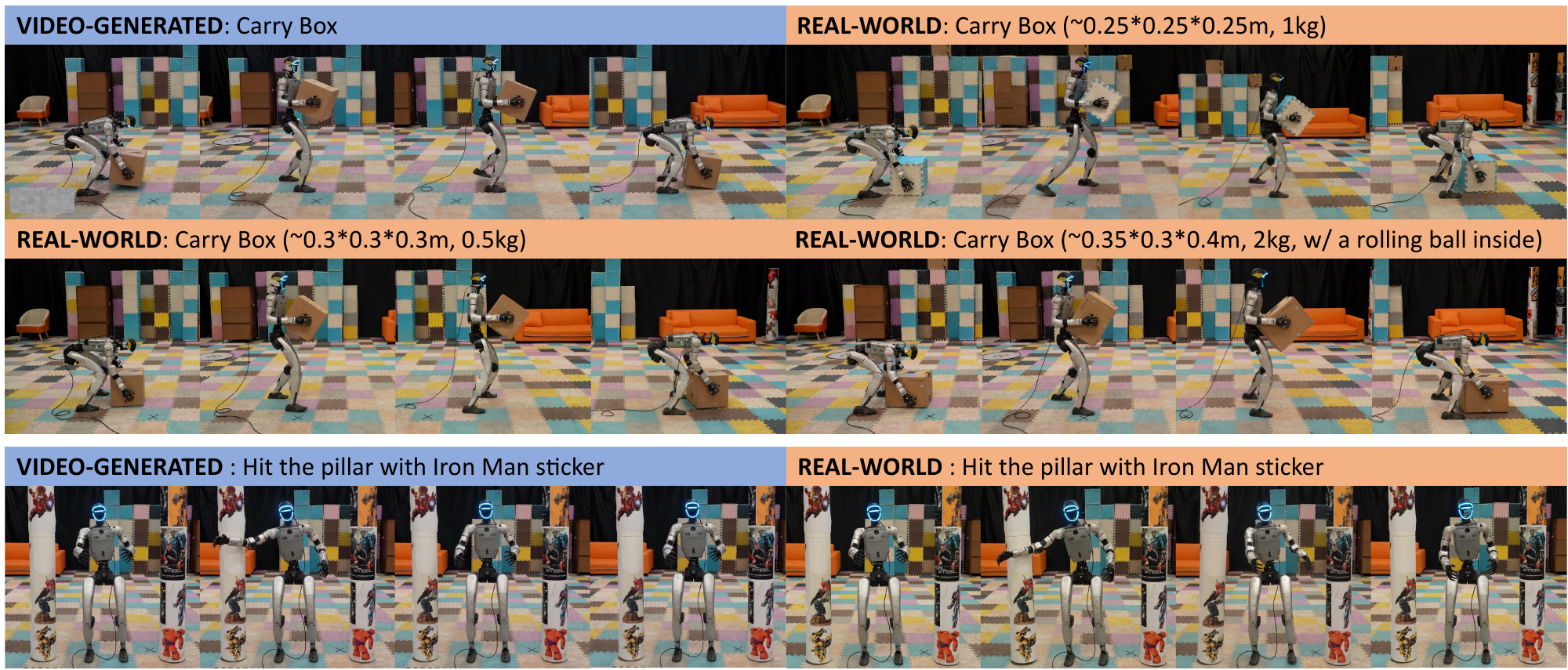}
    \caption{Qualitative results of zero-shot real-world deployment. Time-lapse sequences demonstrate the robot successfully executing diverse physical tasks, including lifting various boxes and performing a semantically rich interaction (hitting an ``Iron Man'' pillar).}
    \label{fig:result_real}
    \vspace{-0.4cm}
\end{figure}

\textbf{Setup. } The real-world deployment pipeline involves three components. First, for video generation, we select Seedance 2.0 Fast \cite{seedance2025}, which exhibits the best instruction-following capabilities and generation consistency. The detailed text prompts used for video generation are provided in Appendix \ref{sec:appendix_prompts}. Second, for point tracking, we utilize SAM3 to segment key parts (the robot's base and hands, and the object) and SpaTrackerV2 to perform 4D point tracking. Finally, we deploy the policy on a Unitree G1 humanoid robot within a motion capture (mocap) system.

\textbf{Zero-Shot Real-World Pipeline. } To answer the third question, we validate the complete zero-shot deployment capabilities of our pipeline in the real world, see \autoref{fig:result_real}. The robot successfully lifts various types of boxes, and by leveraging the generative capabilities of the video model, it can also execute semantically rich interactive tasks, such as hitting an ``Iron Man'' pillar. 

\section{Conclusion and Limitations}
\label{sec:conclusions}

We propose Imagine2Real for zero-shot Humanoid-Object Interaction (HOI). To overcome representation misalignment and retargeting complexity in video-driven methods, we formulate robot and object motions as unified 4D point trajectories and track only sparse critical points. By utilizing the latent space of a Behavior Foundation Model (BFM) as the tracking search domain, Imagine2Real bypasses explicit geometric priors while maintaining natural whole-body gaits. Experiments demonstrate high success rates and robust motion tracking in representative whole-body HOI tasks (e.g., carrying and pushing objects), culminating in successful zero-shot physical deployment on a real humanoid robot.

Our framework currently faces two main limitations. First, real-world deployment relies on motion capture systems, which are prone to marker occlusion during close-contact tasks(e.g., 
pushing boxes). Future work could integrate self-tracking systems using on-device SLAM and multi-camera fusion (e.g., HTC VIVE Tracker). Second, the current pipeline is open-loop. While iteratively prompting the video model provides a pseudo-closed-loop approximation, it is highly inefficient. A true closed-loop system requires jointly fine-tuning the video generator and motion tracker using physical feedback. This remains challenging, as the most capable video models are closed-source APIs, and open-source alternatives (e.g., Wan2.2 \cite{wan2025}) still lack the physical consistency required for complex interactions.


\bibliographystyle{unsrt}
\bibliography{ref}

\clearpage
\appendix
\section*{Appendix}
\section{Training Details}
\label{sec:appendix_training}

In this section, we provide comprehensive training details for the three-stage progressive learning framework introduced in Section \ref{subsec:training}, including the network architectures, hyperparameters, reward functions, and domain randomization strategies. All three policies are trained in the Isaac Gym simulator. 

\subsection{Simulation Setup}
The physics simulation runs at a frequency of 250 Hz ($\Delta t = 0.004$ s). The low-level PD controller updates at 50 Hz, corresponding to a decimation factor of 5. The Keypoints Tracker and Interaction Adaptor operate at a high-level control frequency of 10 Hz ($\Delta t = 0.1$ s), outputting latent residuals and joint residuals, respectively. To robustly simulate the rich variations of the real world and generate diverse interactions, we deploy 8,192 parallel environments per training session.

\subsection{Network Architectures and Optimization}
Table~\ref{tab:hyperparameters} summarizes the detailed network architectures and Proximal Policy Optimization (PPO) hyperparameters for each stage.

\paragraph{BFM Backbone.} 
The BFM backbone is designed to extract a compact 32-dimensional latent space from diverse whole-body motion datasets. The motion encoder compresses local motion sequences into latents using a 6-layer Multi-Layer Perceptron (MLP) with hidden dimensions of 512, followed by a linear layer projecting to 256. The autoregressive prior predictor processes the proprioceptive history using a Gated Recurrent Unit (GRU) with a hidden size of 256, followed by a 6-layer MLP of size 1024. Finally, the decoder (actor) translates the latent representations into target joint positions via an equivalent 6-layer MLP of size 1024. During training, the BFM is optimized with specialized latent consistency losses over 5 epochs and 10 mini-batches per iteration, where the overlap loss coefficient is 0.35, the commitment loss coefficient is 0.25, and the triplet loss coefficient is set to 5.0. 

\paragraph{Keypoints Tracker.}
Building upon the frozen BFM, the Keypoints Tracker learns to output latent residuals to track sparse marker targets. Its actor network fuses the GRU history (dimension 256) with a 3-layer MLP of dimensions $[256, 256, 128]$. The critic network utilizes a 5-layer MLP of dimensions $[512, 512, 512, 512, 256]$. We train this policy with an adaptive learning rate bounded between $1 \times 10^{-5}$ and $1 \times 10^{-4}$ (initialized at $2 \times 10^{-4}$), targeting a KL divergence of 0.05.

\paragraph{Interaction Adaptor.}
The Interaction Adaptor outputs fine-grained joint residuals to complement the Keypoints Tracker. Due to the complexity of object interactions, it employs a slightly larger actor MLP of dimensions $[256, 256, 128]$. It shares the same critic architecture as the Keypoints Tracker but relies on a learnable action noise bounded between $[0.01, 2.0]$ (initialized at 1.0) to encourage enhanced exploration during the contact-rich phases of learning. The target KL divergence is set to 0.02.

\begin{table}[t]
    \centering
    \caption{PPO Hyperparameters and Network Architectures across the three training stages.}
    \vspace{0.1in}
    \small
    \begin{tabular}{lccc}
    \toprule
    \textbf{Hyperparameter} & \textbf{BFM Backbone} & \textbf{Keypoints Tracker} & \textbf{Interaction Adaptor} \\
    \midrule
    \textit{Network Architectures} & & & \\
    Prior/Actor RNN & GRU (256) & GRU (256) & GRU (256) \\
    Actor/Decoder MLP & $[1024]\times 6 \rightarrow 128$ & $[256, 256, 128]$ & $[256, 256, 128]$ \\
    Critic MLP & - & $[512]\times 4 \rightarrow 256$ & $[512]\times 4 \rightarrow 256$ \\
    Latent Dimension & 32 & 32 & - \\
    \midrule
    \textit{Optimization (PPO)} & & & \\
    Learning Rate & Adaptive ($1\times 10^{-4}$) & Adaptive ($2\times 10^{-4}$) & Adaptive ($1\times 10^{-4}$) \\
    Target KL Divergence & 0.02 & 0.05 & 0.02 \\
    Discount Factor $\gamma$ & 0.99 & 0.99 & 0.99 \\
    GAE $\lambda$ & 0.95 & 0.95 & 0.95 \\
    Value Loss Coefficient & 2.0 & 2.0 & 3.0 \\
    Entropy Coefficient & 0.01 & $3\times 10^{-4}$ & $1\times 10^{-3}$ \\
    Clip Parameter & 0.2 & 0.2 & 0.2 \\
    Epochs & 5 & 10 & 5 \\
    Mini-batches & 10 & 8 & 10 \\
    Action Noise & 0.005 (constant) & 0.02 (constant) & 1.0 (learnable) \\
    \bottomrule
    \end{tabular}
    \label{tab:hyperparameters}
\end{table}

\subsection{Reward Functions}
Both the Keypoints Tracker and Interaction Adaptor are trained using straightforward tracking rewards formulated via the exponential kernel function $r = \exp(-\|x - \hat{x}\|^2 / \sigma)$, circumventing the need for manually designed complex heuristics. The detailed reward terms, their corresponding scales, and kernel bandwidth parameters ($\sigma$) are delineated in Table~\ref{tab:rewards}.

For the Keypoints Tracker, tracking precision focuses heavily on the 3D position of the specified marker (e.g., wrist) and the robot base. When advancing to the Interaction Adaptor, the marker tracking bandwidth ($\sigma$) is sharpened from 0.2 to 0.1 to demand tighter precision, and new highly weighted reward terms are introduced to explicitly track the object's 6-DoF pose and interaction geometry (IG). This ensures the robot grasps and manipulates the object strictly as demonstrated. Several regularization penalties, including energy constraints, action smoothness, and feet contact stability, are maintained identically across both stages to preserve natural locomotion priors.

\begin{table}[t]
    \centering
    \caption{Detailed Reward Formulation for the Keypoints Tracker and Interaction Adaptor. Tracking terms utilize an exponential kernel $r = \exp(-\|x - \hat{x}\|^2 / \sigma)$.}
    \vspace{0.1in}
    \small
    \begin{tabular}{lcccc}
    \toprule
    \multirow{2}{*}{\textbf{Reward Term}} & \multicolumn{2}{c}{\textbf{Keypoints Tracker}} & \multicolumn{2}{c}{\textbf{Interaction Adaptor}} \\
    \cmidrule(lr){2-3} \cmidrule(lr){4-5}
    & \textbf{Weight} & \textbf{$\sigma$} & \textbf{Weight} & \textbf{$\sigma$} \\
    \midrule
    \textit{Tracking Rewards} & & & & \\
    Base Position & 4.0 & 0.2 & 4.0 & 0.1 \\
    Base Rotation & 2.0 & 1.2 & 2.0 & 0.6 \\
    Marker Position & 8.0 & 0.2 & 4.0 & 0.1 \\
    Marker Rotation & 2.0 & 1.2 & 2.0 & 1.0 \\
    Joint Position & 4.0 & 20.0 & 4.0 & 10.0 \\
    Joint Velocity & 2.0 & 50.0 & 2.0 & 30.0 \\
    Feet Height & 4.0 & 0.15 & 4.0 & 0.1 \\
    Object Position & - & - & 12.0 & 0.1 \\
    Object Rotation & - & - & 3.0 & 1.0 \\
    \midrule
    \textit{Penalty Rewards} & & & & \\
    Termination & -10.0 & - & -10.0 & - \\
    Joint Torques & -2.0 & - & -2.0 & - \\
    Action Rates & -2.0 & - & -2.0 & - \\
    Action Smoothness & -2.0 & - & -2.0 & - \\
    \bottomrule
    \end{tabular}
    \label{tab:rewards}
\end{table}

\subsection{Domain Randomization}
To facilitate reliable zero-shot sim-to-real transfer, we apply robust domain randomization across the simulation physics parameters, robot configurations, control noises, and external disturbances. During the Interaction Adaptor training, we additionally randomize the target object's physical attributes, including its mass, friction, and scale, to ensure the policy generalizes against perception inaccuracies and physical discrepancies in unstructured real-world deployment. The exhaustive list of randomization parameters and their ranges is provided in Table~\ref{tab:domain_rand}.

\section{Video Generation Prompts}
\label{sec:appendix_prompts}

To ensure the generated videos maintain physical realism and temporal consistency suitable for downstream motion extraction, we append a standardized set of constraints to the task-specific text instructions (e.g., ``The robot walks forward and carries the box''). The complete prompt template is structured as follows:

\vspace{0.1in}
\begin{tcolorbox}[colback=gray!5!white,colframe=gray!75!black,title=\textbf{Video Generation Prompt Template}]
\textbf{[Task Description]}

\textbf{Constraints:}
\begin{itemize}
    \item \textbf{Camera:} Remain fixed and stationary throughout the entire video, with the perspective matching the original image.
    \item \textbf{Robot:} The robot's limbs are rigid bodies, and its motion is restricted to revolute joints without any soft or elastic deformation.
    \item \textbf{Physics:} The robot's motion must be physically accurate, natural, smooth, and realistic.
    \item \textbf{Environment:} Keep the background, floor, and lighting settings in the image unchanged.
\end{itemize}
\end{tcolorbox}

\begin{table}[h]
    \centering
    \caption{Sim-to-Real Domain Randomization Parameters.}
    \vspace{0.1in}
    \small
    \begin{tabular}{ll}
    \toprule
    \textbf{Parameter} & \textbf{Range / Value} \\
    \midrule
    \textit{Robot Dynamics} & \\
    Link Mass & $[0.8, 1.2] \times$ default \\
    Link Center of Mass (CoM) Offset & $[-0.1, 0.1]$ m \\
    Base Payload & $[-0.1, 0.2]$ kg \\
    Link Restitution & $[0.0, 0.5]$ \\
    Link Friction & $[0.3, 1.5]$ \\
    Joint Friction & $[0.5, 2.0]$ N$\cdot$m \\
    Joint Damping & $[0.3, 1.5] \times$ default \\
    Joint Armature & $[1.0, 1.05] \times$ default \\
    \midrule
    \textit{Control \& Perturbations} & \\
    PD Stiffness ($K_p$) & $[0.9, 1.1] \times$ default \\
    PD Damping ($K_d$) & $[0.9, 1.1] \times$ default \\
    Joint Injection Noise & $[-0.01, 0.01]$ rad \\
    Actuation Offset & $[-0.01, 0.01]$ rad \\
    Push Robot Base (Linear Velocity) & $\Delta v_{xy} \le 0.5$ m/s, $\Delta v_z \le 0.2$ m/s (every 1--5s) \\
    Push Robot Base (Angular Velocity) & $\Delta \omega_{xy} \le 0.6$ rad/s, $\Delta \omega_z \le 0.8$ rad/s (every 1--5s) \\
    Push Robot Body (Force) & up to $20$ N for 0.1--0.2s (every 1--5s) \\
    \midrule
    \textit{Object Properties (Interaction Adaptor only)} & \\
    Object Mass & $0.05 \sim 1.5$ kg (base $0.5$ kg $\pm$ noise) \\
    Object CoM Offset & $[-0.05, 0.05]$ m \\
    Object Friction & $[0.1, 1.2]$ \\
    Object Rolling Friction & $[0.0, 0.1]$ \\
    Object Restitution & $[0.0, 0.2]$ \\
    Object Scale & $[0.8, 1.1] \times$ default \\
    Initial Position Offset & up to $0.1$ m \\
    Initial Rotation Offset & up to $5^\circ$ \\
    \bottomrule
    \end{tabular}
    \label{tab:domain_rand}
\end{table}

\end{document}